# Accelerating Ensemble Error Bar Prediction with Single Model Fits

Vidit Agrawal[a], Shixin Zhang[a], Lane E. Schultz[b], and Dane Morgan[b]

[a]University of Wisconsin-Madison, 1210 W. Dayton Street, Madison, WI 53706, USA
[b]University of Wisconsin-Madison, 1500 Engineering Drive, Madison, WI 53706, USA

## 1. Abstract

Ensemble models can be used to estimate prediction uncertainties in machine learning models. However, an ensemble of $N$ models is approximately $N$ times more computationally demanding compared to a single model when it is used for inference. In this work, we explore fitting a single model to predicted ensemble error bar data, which allows us to estimate uncertainties without the need for a full ensemble. Our approach is based on three models: Model A for predictive accuracy, Model $A_E$ for traditional ensemble-based error bar prediction, and Model B, fit to data from Model $A_E$, to be used for predicting the values of $A_E$ but with only one model evaluation. Model B leverages synthetic data augmentation to estimate error bars efficiently. This approach offers a highly flexible method of uncertainty quantification that can approximate that of ensemble methods but only requires a single extra model evaluation over Model A during inference. We assess this approach on a set of problems in materials science.

## 2. Introduction

In the last 10 years there has been a remarkable surge in machine learning for predicting the properties of materials, as which has been explored and documented in several review papers (1–4). Critical for effective use of these models are some forms of uncertainty quantification, which enable researchers to measure the certainty of predictions and provide an assessment of prediction fidelity. In almost all studies aggregate metrics for test prediction errors (e.g., root mean squared or mean absolute error) for models on validation or test data are provided, typically from some form of cross validation. However, it is also very useful to understand the confidence of individual predictions of a target property.

Several uncertainty quantification methods that provide error information for specific predictions of a target value exist (5). One flexible, accurate, and widely used method for error prediction involves the use of ensemble models, which introduce a variance by fitting multiple models to the data, e.g., using bootstrapping dataset (6,7). Building an ensemble of models involves training several models on different samplings of data (bootstrapping), then aggregating predictions from individual models to provide a mean prediction along with their spread. The spread of predictions is a measure of prediction uncertainty. A way to further improve uncertainty estimates involves calibrating the predicted values to align appropriately with observed residuals (4,6). Calibrated uncertainty estimates can be made with ensembles in both classification and regression settings but we focus only on regression applications in this work (6,8–11). Calibrated uncertainties from ensemble models have demonstrated an impressive ability to predict errors in regression models of materials properties, as evidenced by recent research conducted by Palmer et al (6).

Despite the success of ensemble neural network models, an ensemble of N models is expected to be *N* times slower and require *N* times more memory compared to the use of a single model for predictions. This reduced speed can make it less feasible to apply ensemble methods when fast model evaluation is required. In materials science, examples where a fast evaluation is particularly beneficial is during molecular dynamics simulations with machine learning potentials (12) or real-time object detection in images from electron microscopes (13). Issues of speed and memory for ensembles are most likely to be significant for models that are slow and large, a situation that occurs most often with neural network models. Therefore, the focus of this work will be on neural networks, and we will explore relatively small and fast models on modest size data sets to allow for rapid exploration and testing. However, the approaches readily generalize to other types of data and models.

Given the potential issues of speed and memory for ensemble models, it is useful to explore alternative approaches that offer faster model evaluation without compromising predictive accuracy. This paper addresses these concerns by implementing a methodology that avoids the most important computational costs associated with neural network ensemble models while still providing reliable estimates of prediction errors. Specifically, our approach begins with a full ensemble model fit, but then generates training data from that ensemble model and fits a single neural network model to the ensemble error bar predictions. This second model allows uncertainties to be estimated without further using the full ensemble model. Our approach involves sampling the space around the training data to establish a reliable domain for the model for predicting error bars. This approach leverages the advantages of single-model predictions and error bar estimation, which can help add uncertainty quantification to materials property prediction with very modest additional computational and memory costs.

This paper is organized into three sections. Section 3 presents the detailed methods used to implement our methodology of obtaining error bars. Section 4 discusses the learning curves and accuracy of the error bar model observed for different datasets and models. Section 5 summarizes all the statistics. The findings are then examined and discussed in Section 6, which highlights the advantages and potential pitfalls of the method. Section 7 has a summary and conclusions. All our experiments and analyses were conducted in Python using TensorFlow (14), Scikit-Learn (15), and MAST-ML (16).

# 3. Methods

## 3.1 Summary of Approach

This research study utilizes three distinct models for each dataset: Model A, Model $A_E$, and Model B. Model A is a single neural network, developed for predictive accuracy, and trained on the original dataset of features $X_\alpha$ and target $Y_\alpha$. Model $A_E$ is an ensemble of neural networks that are trained and calibrated on the same features $X_\alpha$ and targets $Y_\alpha$. The primary objective of Model $A_E$ is to provide insightful estimations of error bars ($\sigma_A$). Note that by error bars here we mean one standard deviation of the residual distribution, where the residual is the predicted value minus true value. Model B is another single neural network which we train on augmented data $X_\beta$ and targets $Y_\beta$. $Y_\beta$ represents error bar predictions made by Model $A_E$ for the initial and the augmented data. The data augmentation technique involves generating synthetic data points around the original training data point features $X_\alpha$, and Model $A_E$ is leveraged to provide predictions for the newly augmented dataset. By replacing Model $A_E$ with Model B, the complexity of the prediction process is reduced and therefore takes less time and memory. Once the fitting is complete, the combination of Model A and Model B can be used to efficiently provide accurate predictions and error bars. This approach is shown schematically in Figure 1.

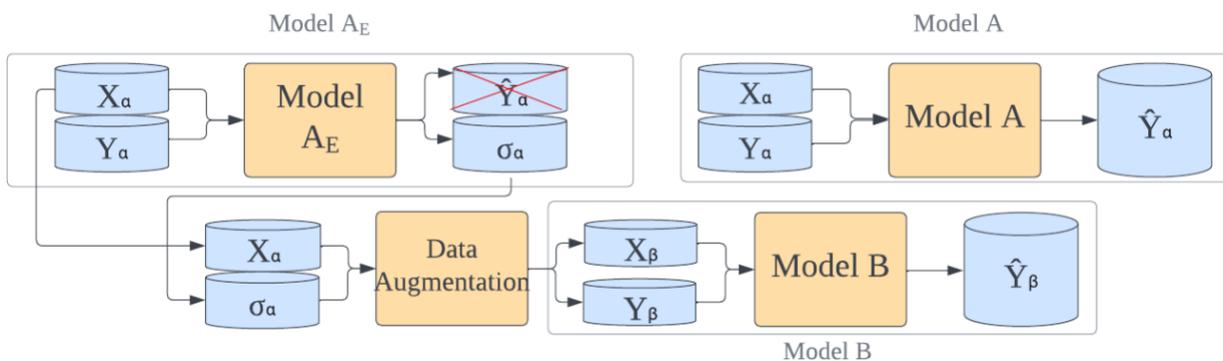

*Figure 1: Flow chart of machine learning system which includes dataset generation, model training, and model prediction. See text for definitions of all the terms.*

## 3.2 Databases and Preprocessing

The proposed methodology was tested on three materials sciences datasets from the literature which we refer to as Diffusion, Perovskite, and Superconductivity. Each makes use of a featurization based on elemental properties and in each case some feature selection was done to keep a manageable number of features. The Diffusion dataset targets are activation energies for impurity diffusion with features generated from elemental data (17,18). The top 20 features for the dataset were identified by a previously conducted study on the dataset (19). The Perovskite dataset targets are perovskite work function values (20) and the top 70 features for the dataset were identified by selecting those with the highest importance from a Random Forest fit to the full dataset. The Superconductivity dataset targets are transition temperatures between the superconducting and normal state, and the features were generated from elemental data. The top 25 features for the dataset were identified by a previously conducted study (21) and used here.

After optimizing the feature set for datasets, we applied MinMaxScaler standardization from the Scikit-Learn library(15) in Python, which rescales values within each column to a range of 0 to 1. These preprocessing steps were consistently implemented across all stages of the model development processes, including both Model $A_E$ and Model B. Feature selection is particularly important for the approach taken in this work as Model B is trained on augmented data in a volume of feature space whose size scales with the number of features. Therefore, fewer features are likely to yield a more accurate Model B. Feature standardization is also important as the volume in feature space used for data augmentation is controlled by a unitless length scale that start small ($10^{-3}$) and grows to a significant value (0.5) relative to the value one. Therefore, this approach can only be expected to yield results like those shown here for scaled features.

## 3.3   Single Neural Network for High Predictive Accuracy (Model A)

Model A in our study is a single neural network which has been trained on the initial dataset $\{\alpha\}=(X_\alpha,Y_\alpha)$ to provide high accuracy for $Y_\alpha$. The model has an architecture of two linear layers with 2048 neurons and Rectified Linear Unit (ReLU) activation function. The loss function of this neural network uses mean squared error and uses the Adam optimizer. The network was trained for 100 epochs. Despite observations indicating the convergence is typically achieved by the 50th epoch, the decision to extend the training to 100 epochs was made to ensure the robustness of the model.

## 3.4   Ensemble Models for Error Bars (Model $A_E$)

Model $A_E$ in our study is an ensemble of 20 Fully Connected Neural Network models constructed on bootstrapped subsets of data set $\{\alpha\}=(X_\alpha,Y_\alpha)$. Each network has the same architecture and training approach as that for Model A in section 3.3. The calibrated spread in values from this ensemble are used to predict error bars ($\sigma_A$). These error bars were calibrated for better accuracy using the method mentioned in Palmer, et al (6).

## 3.5 Data Generation for Model B fitting

Model B is fit to data set $\{\beta\} = (X_\beta, Y_\beta)$, which consists of feature values and their associated error bars as predicted by model $A_E$. Since the training data for Model B are error bars that can be generated easily from model $A_E$ it is straightforward to set the data for fitting model B to give it desired accuracy and domain, at least within practical computational limitations. There is no forced restriction on what data should be generated and used for fitting model B so a choice must be made. A natural set of feature points to use are the original set of feature points in the database $\{\alpha\}$, $X_\alpha$, which were used to fit the model A and $A_E$ (although with different targets than will be used for model B). These points are likely to be near points of interest for future predictions. So, we will consider this our initial database and denote it $\{\beta_0\}$. Then we augment this database with datapoint that are nearby in feature space to this starting database. Specifically, we generate data by randomly sampling in feature space near the values in $\{\beta_0\}$ using the following approach. First, assumed we have scaled the features, so each component for the features in $\{\beta_0\}$ ranges from 0 to 1. If the scaled features of a datapoint in $\{\beta_0\}$ is denoted as **x** then we sample randomly from the hypercube of $\mathbf{x} \pm s$, where $s$ is a scale factor. In other words, the sampled points in feature space can range from $x_i - s$ to $x_i + s$ for each component of the feature vector. The scale factor $s$ is allowed to vary from 0.001 to 0.5, where larger values sample a larger volume of feature space. All these sampled feature spaces include the original set $\{\beta_0\}$ and will have increased sample sizes for all the different augmented datasets. In cases when the sampling leads to values of components outside of the interval [0,1] the value is taken as the maximum or the minimum value of the interval [0,1], ensuring it does not exceed observed maximum or minimum values in the original dataset. By employing this randomization technique, possibly truncated hypercubes are generated around each data point in the original database, providing a relevant volume in feature space for sampling to achieve a useful domain over which we will train model B to learn accurate error bars.

The data generation scheme chosen here is certainly not unique and other approaches could be used. A particular limitation of the present approach is that no effort is made to assure that the data points are chemically feasible in the context of material science. It is likely that during the augmentation process data points are created that do not conform to the constraints and properties of real-world materials. Physically constrained data generation could provide sampling in a much more constrained region of feature space, potentially leading to more accurate model B's with less training data.

## 3.6 Single Neural Network Error Bar Generation (Model B)

After data augmentation, the augmented dataset $X_\beta$ is created, and Model $A_E$ is used to obtain the error bars for these augmented datapoints. These error bars become the target variable $Y_\beta$ for Model B. Model B is trained using the augmented dataset $\{\beta\} = (X_\beta, Y_\beta)$ to learn from the augmented data and generate accurate error bars in subsequent predictions. To ensure standardized input values, the augmented dataset $X_\beta$ undergoes MinMaxScaler standardization as neural networks require standardized features. The predictions generated by Model B, $Y_\beta$, serve as the estimated error bars. In using our approach one can replace Model $A_E$ with Model B for error bar predictions, reducing time and memory requirements.

## 3.7 Use Case of the Workflow

To effectively utilize our technique in practical applications one would take the following steps. First, Model A is fit, which can then be employed in future predictions. Then model $A_E$ is fit, synthetic data $X_\beta$ is generated, and model $A_E$ is used to predict error bars ($\sigma_A$). Then model B is fit to the data $X_\beta$ and error bars ($Y_\beta = \sigma_A$) so the model B can predict approximate error bars for model A ($\hat{Y}_\beta$) in the future. In applications to predict a materials property with error bars, Model A is used to predict values $\hat{Y}_A$ and Model B is used to predict approximate model A errors bars ($\hat{Y}_\beta$), eliminating the need for Model $A_E$ in evaluation.

# 4 Results

The critical question for the approached outlined in Section 3 (Methods) is whether model B can represent the error bars accurately. If this representation can be done accurately then the proposed approach can be a practical and general method to avoid slowing down model prediction when using ensemble methods for error bar prediction. Therefore, in this section our focus is on the analysis of the accuracy of Model B. We focus on learning curves that evaluate the 5-fold cross validation (CV) results of Model B on each augmented dataset. The graphs show the normalized CV root-mean-squared-error (RMSE), which is the RMSE obtained for out of bag data during CV, scaled using standard deviation of the target variable ($Y_\beta$), which we denote by "Sigma". We consider training datasets augmented up to $10^6$ data points. We used normalized CV-RMSE as this is easy to interpret, since a value of one is equivalent to that one would obtain by simply guessing the mean of the target data. These learning curves provide insights into the impact of the amount of training data and the varying scale factor. The results demonstrate that the proposed methodology for estimating error bars provides accurate models when applied to small to modest scale factors for manageable database sizes, However, larger scale factors lead to reduced accuracy, likely above what would be considered useful. We also evaluate the performance of this method across three distinct datasets to demonstrate the generality of the results. Further analysis, with multiple statistics for a wider set of models, is provided in the Supplemental Information (see section 12 for details).

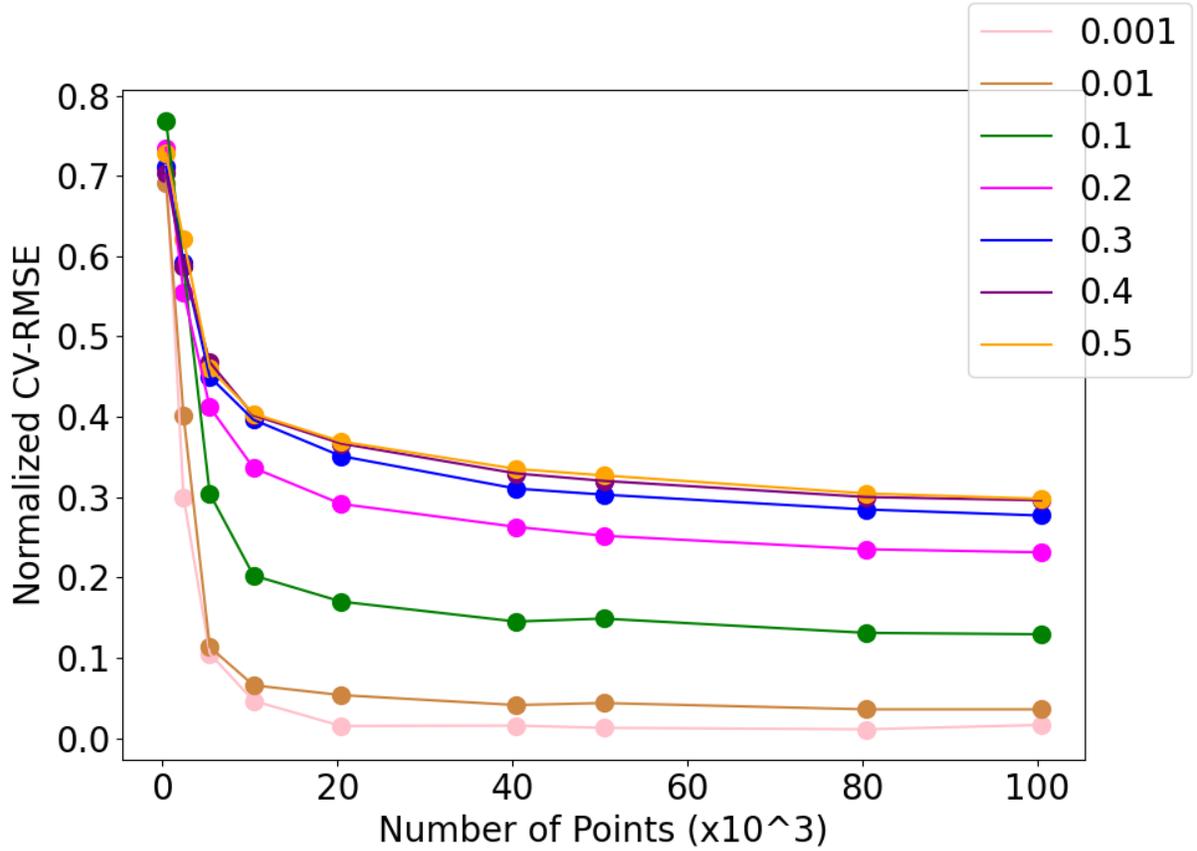

*Figure 2(a): Relationship between the increasing the number of points present in the augmented dataset (Diffusion) and the decreasing normalized CV-RMSE of predicted $\hat{Y}_\beta$ (RMSE/Sigma) for Diffusion dataset trained on Neural Networks. The legends show the varying scale factors for the different data augmentations. (See Data Generation section 3.5 for more information)*

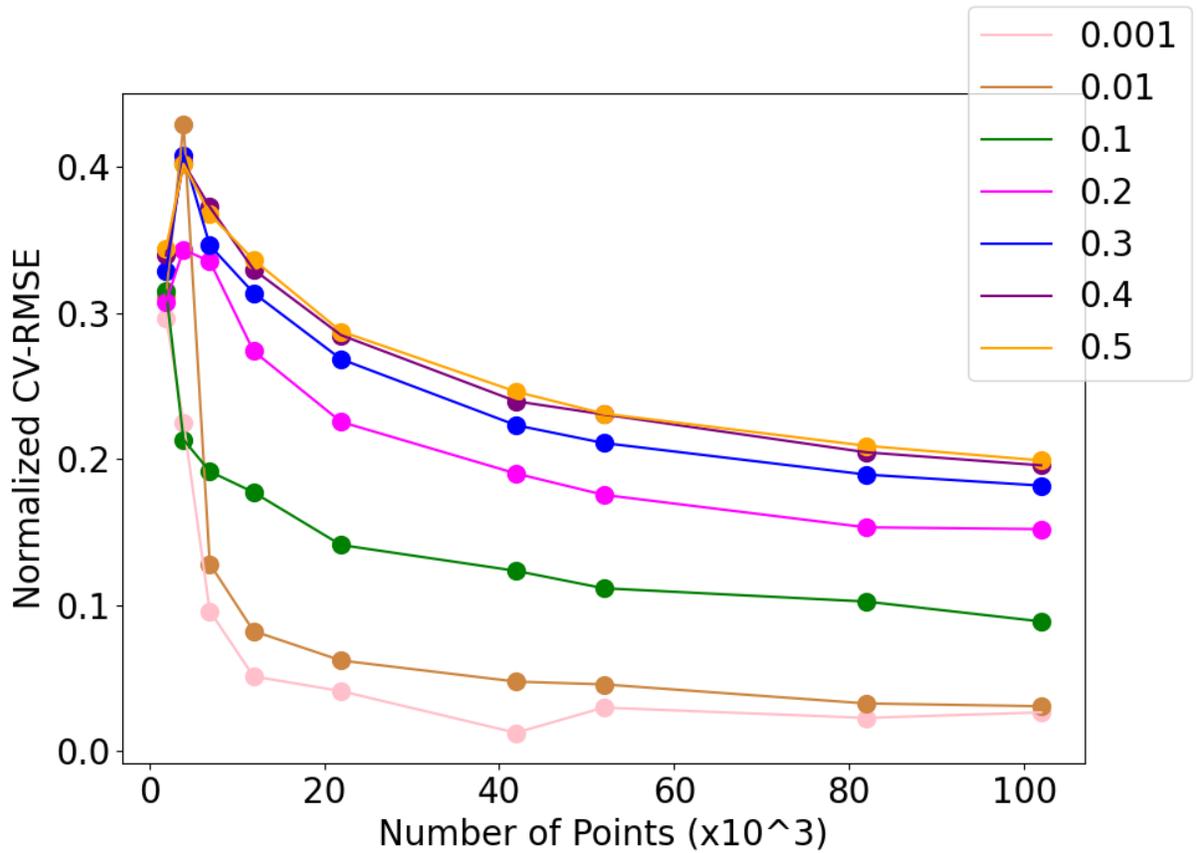

*Figure 2(b): Relationship between the increasing the number of points present in the augmented dataset (Perovskite) and the decreasing normalized CV-RMSE of predicted $\hat{Y}_\beta$ (RMSE/Sigma) for Perovskite dataset trained on Neural Networks. The legends show the varying scale factors for the different data augmentations. (See Data Augmentation section for more information)*

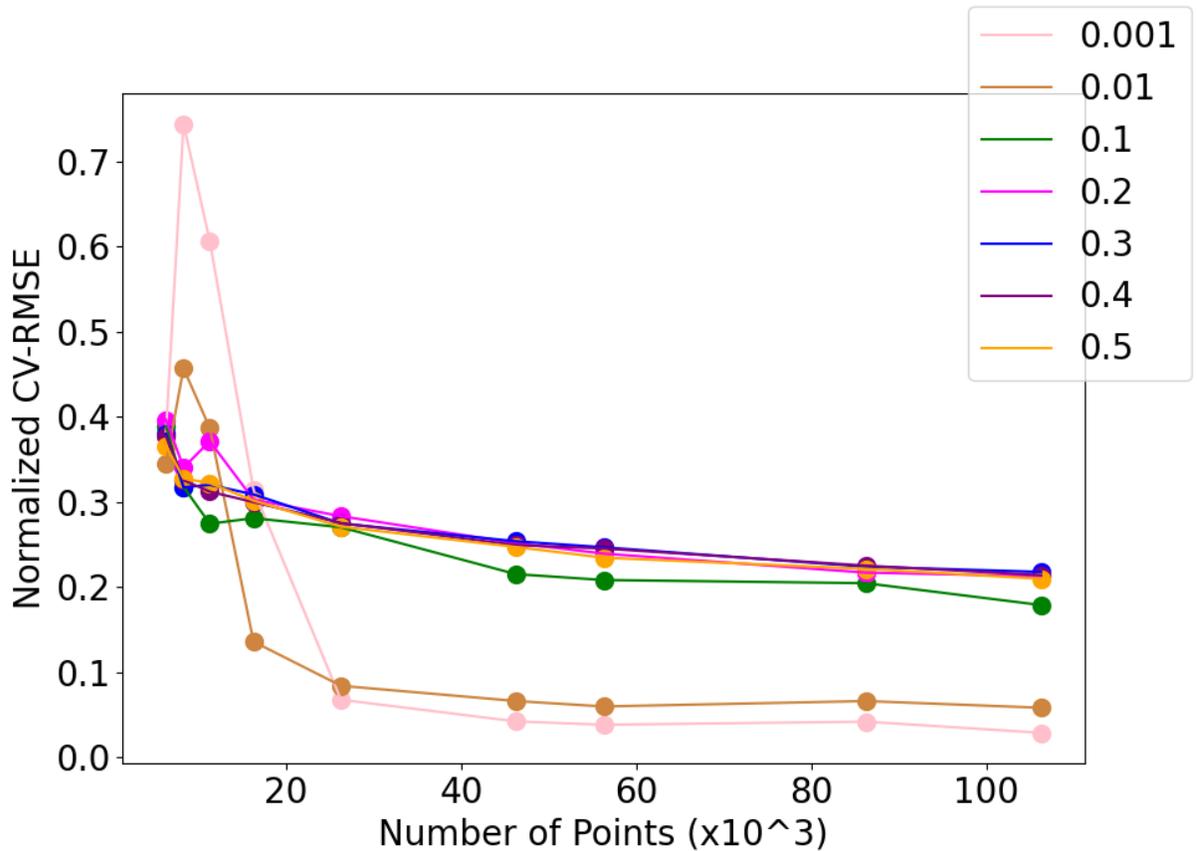

*Figure 2(c): Relationship between the increasing the number of points present in the augmented dataset (Superconductivity) and the decreasing normalized CV-RMSE of predicted $\hat{Y}_\beta$ (RMSE/Sigma) for Superconductivity dataset trained on Neural Networks. The legends show the varying scale factors for the different data augmentations. (See Data Augmentation section for more information)*

As expected, when the number of points in the dataset increases, the normalized CV-RMSE of Model B decreases, with a decreasing rate of reduction as the more points are added. Also as expected, the normalized CV-RMSE values are smaller and converge to faster for smaller scale factors, consistent with their being associated with a smaller feature space volume that must be modeled.

We have also tried replacing Model Bs with a random forest regression model and we obtain similar results (See Supplementary Information (Model B)). We also tested different Models A's, including random forest regression, K-nearest neighbors, and neural networks with Model B's being a random forest model, and generally found very similar trends to those shown here. These results are also shown in the Supplement Information. Overall, the similarity of the results across data sets and different Model A's and B's suggests that our approach and qualitative results are general.

# 5 Statistics Table

*Table 1: Test Statistics for Diffusion, Perovskite, and Superconductivity datasets. All errors on out of bag data from 5-fold CV. The three values shown in each table cell represent fit with original number of points / fit with max number of points (100,000) / Best possible fit out of all the different number of points. The cells marked with "*" represent the ones where the max number of points fit is not the best fit.*

| Dataset | Scale Factor | Standard Deviation (σ) | MAE | R2 | Normalized CV-RMSE | RMSE |
|---|---|---|---|---|---|---|
| Diffusion | 0.001 | 0.03/0.03/0.03 | 0.01/0.0/0.0 | 0.5/1.0/1.0 | 0.71/0.01/0.01 | 0.02/0.0/0.0 |
| | 0.01 | 0.03/0.03/0.03 | 0.01/0.0/0.0 | 0.49/1.0/1.0 | 0.71/0.03/0.03 | 0.02/0.0/0.0 |
| | 0.1 | 0.03/0.03/0.03 | 0.02/0.01/0.01 | 0.48/0.92/0.92 | 0.72/0.29/0.29 | 0.02/0.01/0.01 |
| | 0.2 | 0.03/0.03/0.03 | 0.01/0.01/0.01 | 0.5/0.75/0.75 | 0.71/0.5/0.5 | 0.02/0.01/0.01 |
| | 0.3 | 0.03/0.03/0.03 | 0.01/0.01/0.01 | 0.47/0.64/0.64 | 0.73/0.6/0.6 | 0.02/0.02/0.02 |
| | 0.4 | 0.03/0.04/0.04 | 0.02/0.02/0.02 | 0.47/0.61/0.61 | 0.73/0.63/0.63 | 0.02/0.02/0.02 |
| | 0.5 | 0.03/0.04/0.04 | 0.01/0.02/0.02 | 0.49/0.6/0.6 | 0.71/0.64/0.64 | 0.02/0.03/0.03 |
| Perovskite | 0.001 | 20.11/19.92/19.92 | 4.62/0.06/0.06 | 0.53/1.0/1.0 | 0.68/0.01/0.01 | 13.77/0.24/0.24 |
| | 0.01 | 20.11/19.82/19.82 | 4.52/0.48/0.48 | 0.7/1.0/1.0 | 0.55/0.05/0.05 | 11.08/1.0/1.0 |
| | 0.1 | 20.11/20.18/20.18 | 4.47/4.18/4.18 | 0.68/0.91/0.91 | 0.56/0.3/0.3 | 11.31/6.08/6.08 |
| | 0.2 | 20.11/21.66/22.2 | 4.53/7.99/8.07 | 0.66/0.73/0.73 | 0.59/0.52/0.52* | 11.79/11.31/11.5 |
| | 0.3 | 20.11/25.08/20.11* | 4.52/11.5/4.52* | 0.64/0.61/0.64 | 0.6/0.63/0.6* | 12.08/15.76/12.08* |
| | 0.4 | 20.11/29.2/20.11* | 4.46/14.77/4.46* | 0.69/0.54/0.69 | 0.56/0.68/0.56* | 11.23/19.89/11.23* |
| | 0.5 | 20.11/33.66/20.11* | 4.38/17.74/4.38* | 0.71/0.5/0.71 | 0.54/0.71/0.54* | 10.84/23.8/10.84* |
| Superconductivity | 0.001 | 10.87/10.8/10.8 | 0.93/0.08/0.08 | 0.84/1.0/1.0 | 0.4/0.02/0.02 | 4.3/0.19/0.19 |
| | 0.01 | 10.87/10.78/10.78 | 0.9/0.55/0.55 | 0.86/0.99/0.99 | 0.37/0.09/0.09 | 4.01/0.94/0.94 |
| | 0.1 | 10.87/12.62/10.87* | 0.92/3.67/0.92* | 0.86/0.81/0.86 | 0.37/0.44/0.37* | 4.01/5.5/4.01* |
| | 0.2 | 10.87/17.51/10.87* | 0.91/6.41/0.91* | 0.86/0.72/0.86 | 0.38/0.53/0.38* | 4.09/9.31/4.09* |
| | 0.3 | 10.87/22.23/10.87* | 0.92/8.61/0.92* | 0.85/0.69/0.85 | 0.38/0.56/0.38* | 4.18/12.4/4.18* |
| | 0.4 | 10.87/26.35/10.87* | 0.89/10.61/0.89* | 0.87/0.66/0.87 | 0.37/0.58/0.37* | 3.98/15.28/3.98* |
| | 0.5 | 10.87/30.1/10.87* | 0.93/12.29/0.93* | 0.85/0.65/0.85 | 0.39/0.59/0.39* | 4.27/17.84/4.27* |

# 6 Discussion

It is somewhat unclear for what values one would say we have a successful model B. In particularly, different users might need different scale values for their domain or different levels of accuracy, as captured by the normalized CV-RMSE. In addition, users may have different maximal values of number of points they are willing to use in training Model B. Barring numerical issues, the RMSE will go to zero with enough points, as the sampling will eventually cover the entire feature space with a very fine mesh requiring only limited interpolation.

However, users will typically have a finite computational budget for training Model B, so there will be some maximum number of points that can be practically treated. To have some practical guidance in this work we will consider a scale factor of 0.1 usefully large, and normalized CV-RMSE < 0.2 usefully small, and $10^6$ points as a practical maximum for training data size for model B. Figure 2 shows that across all three datasets model B can reproduce the error bars with a normalized CV-RMSE of ≤ 0.1 (0.2) for a scale factor below $10^{-2}$ ($10^{-1}$) within $10^6$ training points. We believe that this demonstrates that a practically accurate model B for error bars of model A can be achieved for a reasonable volume of feature space with a practical number of training points.

However, Figure 2 also shows that there are limitations when the scale factor is as high as even 0.2. Despite increasing the number of data points, Model B struggles to fit the data effectively, resulting in a normalized CV-RMSE ranging from 0.18 to 0.25 even at $10^6$ training points. Larger scale factors generally have even larger normalized CV-RMSE. The elevated normalized CV-RMSE with scale factor is a consequence of the enlarged feature space volume being explored, which results in much more limited sampling and possibly much greater variation of the target, thereby challenging Model B's ability to achieve accurate prediction.

In Figure 2 we can observe that for certain scale factors there is a significant increase in the normalized CV-RMSE after the sampling of the first point, which represents the original set of points. This increase can be attributed to the training data changing dramatically in going from the original data to a database primarily sampled from a high-dimensional feature space. This change in the nature of the data can be seen in the CV-RMSE and Sigma learning curves present in the Supplemental Information. Once the database becomes dominated by the number of sampled points the normalized CV-RMSE trend is always decreasing, except due to small fluctuations from the stochastic nature of the neural networks.

# 7 Summary and Conclusions

In this research, a novel approach is introduced to efficiently estimate error bars ($\sigma_A$) in machine learning models. Our method combines three models: Model A for predictive accuracy, Model $A_E$ for ensemble-based error bar prediction, and Model B, which efficiently estimates error bars by fitting to Model $A_E$. Model B, a single model trained on augmented data, replaces the ensemble Model $A_E$, reducing computational demands during inference while maintaining predictive accuracy. This approach offers a flexible and efficient means of uncertainty quantification. We demonstrate that this approach is practical on examples from the domain of materials science.

The Model B approach circumvents the use of ensembles for uncertainty quantification with single, less computationally intensive models. The approach was shown to be effective with three datasets (denoted Diffusion, Perovskite, Superconductivity) and with the use of a simple neural network as Model B (additional machine learning approaches were also shown to work well in the Supplement Information). Accurate and reliable error bar predictions were observed for cases with small to modest sampling areas around the training data, but the technique's effectiveness diminishes as the sampled area exceeds hypercubes of sides more than +/- 0.2 around the original scaled data points.

This work demonstrates a practical approach to achieve error bar estimation with accuracy approaching that of ensemble methods using just a single model for the error predictions. Such a method can enhance the speed and reduce the memory required by ensemble methods and thereby support greater use of uncertainty quantification in machine learning.

# 8 Data and Code Availability

We have made all the original databases used in this work available in the GitHub repository. We have made the python code used to perform all the calculations and generate all figures publicly available on GitHub in the same repository as the data described above (https://github.com/uw-cmg/material_error_bar_predictions). We also have published all the model splits; pickled model files and data of the runs can be found in Zenodo (10.5281/zenodo.10934013).

# 9 Acknowledgements

The National Science Foundation provided support for Vidit Agrawal, Shixin Zhang, Lane E. Schultz, and Dane Morgan by the NSF Collaborative Research: Framework: Machine Learning Materials Innovation Infrastructure (award number 1931306). Additionally, Vidit Agrawal was provided financial support by University of Wisconsin-Madison's Sophomore Research Fellowship. We thank Ryan Jacobs for many helpful discussions.

# 10 Author Contributions

Vidit Agrawal, Shixin Zhang, Lane E. Schultz, and Dane Morgan developed the accelerated error bar method presented. Vidit Agrawal, Shixin Zhang, and Lane E. Schultz wrote code and analyzed output data. Vidit Agrawal wrote the first draft of the manuscript. Vidit Agrawal, Lane E. Schultz, and Dane Morgan edited and contributed to the later drafts of the manuscript.

# 11 Competing Interests

The authors declare no competing interests.

## 12 Supplemental

The supplemental information for this paper is a digital supplement with a readme HTML file which has all the links to the different plots accessible related to both Model $A_E$ and Model B. It also has all the test statistic data, and different parity plots.

## 13 Disclosure of generative AI use

During the preparation of this work the author(s) used ChatGPT in order to better refine the language. After using this tool/service, the author(s) reviewed and edited the content as needed and take(s) full responsibility for the content of the publication.